\documentclass[sigconf, screen]{acmart}
\renewcommand\footnotetextcopyrightpermission[1]{} % hide footnote in the first page
\AtBeginDocument{%
  }

\setcopyright{acmlicensed}
\copyrightyear{2018}
\acmYear{2018}
\acmDOI{XXXXXXX.XXXXXXX}
\acmConference[Conference acronym 'XX]{Make sure to enter the correct
  conference title from your rights confirmation email}{June 03--05,
  2018}{Woodstock, NY}
\acmISBN{978-1-4503-XXXX-X/2018/06}

\usepackage{enumitem}
\usepackage{bbm}
\usepackage{multirow}
\usepackage{subcaption}
\usepackage{graphicx}
\DeclareMathOperator*{\argmax}{arg\,max}

\newcommand\independent{\protect\mathpalette{\protect\independenT}{\perp}}
\def\independenT#1#2{\mathrel{\rlap{$#1#2$}\mkern2mu{#1#2}}}

\begin{document}

\title{FairITR: Fairness-Aware Learning of Multicategory Individualized Treatment Rules}

\author{Zeyu Chen}
\correspondingauthor
\affiliation{%
  \institution{University of Delaware}
  \city{Newark}
  \state{DE}
  \country{USA}
}
\email{chenze@udel.edu}

\author{Wei Qian}
\affiliation{%
  \institution{University of Delaware}
  \city{Newark}
  \state{DE}
  \country{USA}
}
\email{weiqian@udel.edu}

\author{Bintong Chen}
\affiliation{%
  \institution{University of Delaware}
  \city{Newark}
  \state{DE}
  \country{USA}
}
\email{bchen@udel.edu}


\begin{abstract}
Individualized treatment rules (ITRs) recommend treatment decisions based on individual characteristics and can improve treatment outcomes beyond traditional one-size-fits-all strategies. However, similar to other commonly used algorithmic decision-making methods, ITRs without considering model fairness may result in unfair decisions across groups defined by sensitive attributes, such as race or gender.
Although extensive methods have been developed for ITR estimation, most focus solely on optimizing treatment outcomes, while research on fairness-aware ITRs remains limited. Moreover, the practical scenario of multicategory treatments is rarely considered in existing fair ITR studies.
To address these gaps, we introduce a fairness notion for multicategory ITRs that requires treatment decisions to be independent of sensitive attributes, and then formulate the corresponding problem of estimating fair ITRs.
To solve this problem, we further propose FairITR, a fairness-aware learning framework that jointly optimizes treatment outcomes and fairness. FairITR builds on angle-based classification for multicategory ITR learning and integrates adversarial learning with covariance-based regularization to mitigate unfairness in treatment decisions. A theoretical justification for the fairness properties induced by the adversarial debiasing mechanism is also provided.
Extensive experiments on real-world datasets show that FairITR substantially reduces unfairness in ITRs while maintaining competitive treatment values relative to existing baselines.
\end{abstract}





\maketitle

\section{Introduction}
Individualized treatment rules (ITRs) are statistical machine learning methods that tailor treatment recommendations to individuals by mapping individual characteristics to treatment decisions. Here, ``treatment'' is not limited to its conventional medical meaning but broadly refers to interventions or decisions across a wide range of fields. With the increasing availability of individual-level data, ITRs have attracted growing attention in recent years and found applications in diverse domains, including precision medicine \cite{chen2025federated}, social policy interventions \cite{athey2021policy}, and recommender systems \cite{ma2020off}.

%There has been substantial methodological research on the estimation of ITRs, with existing methods generally falling into two categories: indirect methods and direct methods. Indirect methods first estimate intermediate quantities that characterize the optimal treatment decision and then derive the ITR from these estimates. Representative approaches include Q-learning \cite{watkins1992q, murphy2005generalization, qian2011performance, goldberg2012q, song2015penalized} and A-learning \cite{robins2000marginal, murphy2003optimal, shi2018high}. In contrast, direct methods, also known as policy search methods, directly estimate treatment rules by maximizing the expected outcome over a prespecified class of candidate rules. A notable approach in this category is outcome weighted learning \cite{zhao2012estimating}, followed by several extensions that improve robustness \cite{zhang2012robust, liu2018augmented}, accommodate negative outcome values \cite{zhou2017residual}, and address the non-differentiability of the loss function \cite{jiang2019entropy}. These methods primarily consider binary treatments, while many real-world problems deal with multicategory treatments. Subsequent studies have been effectively developed for multicategory settings \cite{zhang2020multicategory, huang2019multicategory, lee2024effective}.

By assigning treatment decisions based on individual characteristics, ITRs depart from the traditional one-size-fits-all approach and can improve overall outcomes with more tailored interventions. Examples include improved treatment outcomes through personalized interventions in genomic medicine \cite{khan2025genomic} and increased profitability through personalized pricing in retail \cite{chen2026graph}. However, ITRs also raise serious fairness concerns because their decisions are generated algorithmically and may systematically discriminate against certain protected groups defined by sensitive attributes (e.g., race or gender) \cite{wang2022brief, pessach2023algorithmic}. For example, some experimental hiring tools designed to automatically filter job applicants were reported to be biased against women \cite{dastin2022amazon,tyagi2024enhancing}. Similarly, the COMPAS recidivism risk assessment tool was shown to produce disproportionately high risk assessments for certain population groups \cite{angwin2016report}. These fairness concerns are particularly important in high-stakes settings, where individualized decisions can have substantial societal consequences and potentially conflict with relevant regulations (e.g.,  Equal Credit Opportunity Act for credit approval; Fair Housing Act for housing service decisions).

Despite the importance of fair individualized decision-making, research on fairness-aware ITRs that explicitly incorporate fairness considerations into ITR estimation remains limited. Moreover, fairness in ITRs can be defined in different ways depending on the application context. For example, \citet{fang2023fairness} considers fairness from the perspective of protecting individuals with relatively poor outcomes. Their approach requires a lower quantile of the outcome distribution under an ITR to exceed a prespecified threshold. On the other hand, in the algorithmic fairness literature, fairness is more commonly characterized by disparities in decision outcomes with respect to sensitive attributes \cite{pessach2022review, wan2023processing, kamishima2012fairness, calders2010three}. Given its broader use, the fairness criterion in our work is closely aligned with this definition. Along this line, \citet{frauen2024fair} introduce action fairness, which requires treatment decisions to be independent of sensitive attributes, and further propose a neural network-based framework for learning fair ITRs under this criterion. However, their method is limited to binary treatments and optimizes fairness and outcome objectives in two separate stages. Such a binary-treatment restriction can be limiting in practice, as many real-world applications involve multiple treatment options. For conventional ITR methods, a simple strategy to address fairness concerns is fairness through unawareness, which excludes sensitive attributes from the model input \cite{mehrabi2021survey}. Although this strategy is intuitively appealing, it is generally insufficient because other covariates may remain correlated with sensitive attributes, serve as proxy features, and finally lead to unfair decisions \cite{cornacchia2023auditing, gajane2017formalizing}. These limitations motivate a fairness-aware ITR learning framework that optimizes treatment outcomes and fairness jointly in the multicategory treatment setting.

Accordingly, in this work, we first introduce a fairness criterion for multicategory ITRs that promotes fair decision-making across groups defined by sensitive attributes. Based on this criterion, we then formulate fair ITR estimation as a constrained optimization problem that maximizes expected treatment outcomes subject to fairness constraints.
Considering the promising empirical performance of direct ITR learning methods \cite{jiang2019entropy, zhao2012estimating, zhang2012robust, zhou2017residual}, which directly search for optimal treatment rules within a prespecified class of candidate rules, we solve this problem following the direct learning paradigm. Specifically, we propose a novel fairness-aware ITR learning framework, \textbf{FairITR}, for estimating neural network-based ITRs in the multicategory treatment setting. FairITR builds on the angle-based classification framework to handle multiple treatment options. To reduce the dependence of treatment decisions on sensitive attributes, we incorporate an adversarial learning module that debiases the hidden representations of the neural network. We also include an additional fairness regularization term on the network output to further stabilize fairness performance. Together, these components enable FairITR to jointly optimize treatment outcomes and fairness. Our main contributions can be summarized as follows:
\begin{itemize}[topsep=10pt, itemsep=5pt]
    \item This work introduces a fairness notion for multicategory ITRs and formulates the corresponding problem of estimating fair ITRs. To the best of our knowledge, it is the first to study fair ITRs in the multicategory treatment setting.
    \item We propose FairITR, a novel fairness-aware ITR learning framework for neural network-based multicategory ITRs that jointly optimizes treatment outcomes and fairness.
    \item Extensive experiments including three real-world datasets demonstrate that FairITR effectively mitigates unfairness in treatment decisions while maintaining competitive treatment values compared with baseline methods.
\end{itemize}

\section{Related Work}
In this section, we review two research streams that are most closely related to our work: individualized treatment rules and fairness in machine learning. 
We focus on how our work connects to prior studies and where it differs from them.

\subsection{Individualized Treatment Rules (ITRs)}
ITRs map individual characteristics to treatment recommendations with the goal of maximizing the expected treatment outcome \cite{qian2011performance, zhao2019efficient, chen2018estimating}. ITRs can be learned from randomized trial data or observational study data, although the latter requires certain causal assumptions to identify the expected outcome \cite{schulte2015q}.

There has been substantial research on the estimation of ITRs, with existing methods generally falling into two categories: indirect and direct methods. Indirect methods typically follow a two-step procedure: they first estimate an outcome regression model and then derive the optimal ITR from this estimated model. Q-learning \cite{watkins1992q, murphy2005generalization, qian2011performance, goldberg2012q, song2015penalized} and A-learning \cite{robins2000marginal, murphy2003optimal, shi2018high} are two prominent approaches in this class. Q-learning estimates the conditional mean of the outcome given covariates under each treatment and assigns the treatment that has the largest estimated value. By comparison, A-learning models the conditional outcome contrast between treatments given covariates, which is sufficient for determining the optimal treatment decision. This can provide greater robustness to misspecification of the baseline outcome model \cite{schulte2015q}.
In contrast, direct methods, also known as policy search methods, avoid modeling the full outcome regression and instead directly optimize the treatment rule itself. A notable approach in this category is outcome weighted learning (OWL) \cite{zhao2012estimating}; it formulates the ITR estimation as a weighted classification problem by showing that maximizing the expected treatment outcome is equivalent to minimizing a weighted classification loss. Building on OWL, subsequent studies extend this framework in several directions. Augmented OWL methods \cite{zhang2012robust, liu2018augmented} improve its robustness to model misspecification, residual weighted learning (RWL) \cite{zhou2017residual} better handles negative outcome values and improves its finite-sample performance, and entropy learning \cite{jiang2019entropy} replaces the non-differentiable OWL loss with a smooth alternative to facilitate statistical inference for parameters in ITRs. These methods primarily consider binary treatments, while many real-world problems deal with multicategory treatments. Therefore, further developments address the nontrivial challenge of generalizing direct methods from binary to multicategory treatments \cite{zhang2020multicategory, huang2019multicategory, lee2024effective}. Our proposed framework follows this direct learning paradigm in the multicategory treatment setting, although existing multicategory ITR methods do not explicitly take fairness into consideration. 

%Recent studies highlight that ITRs can suffer from fairness issues, motivating efforts to develop fair ITR methods that mitigate unfairness in treatment decisions \cite{fang2023fairness, frauen2024fair, wang2025advancing}. \citet{fang2023fairness} address this issue by augmenting the treatment-outcome objective with a fairness regularization term that controls the lower quantile of the outcome distribution. \citet{frauen2024fair} propose a neural network-based approach that first applies adversarial debiasing to learn fair covariate representations and then estimates the ITR from these representations. This two-stage design separates fairness mitigation from treatment-outcome optimization. \citet{wang2025advancing} further study fair ITRs for survival outcomes under censoring.

%Existing multicategory ITR methods do not explicitly consider fairness, while existing fair ITR approaches remain limited to binary treatments, with some conducting fairness mitigation and outcome optimization in separate stages. Our work differs by jointly optimizing treatment outcomes and fairness in the multicategory setting.

\subsection{Fairness in Machine Learning}
Modern machine learning algorithms have been widely deployed across many domains and achieved substantial success. At the same time, growing evidence shows that these systems can produce unfair outcomes for individuals based on sensitive attributes \cite{mehrabi2021survey}. One major source of such unfairness is that machine learning models may inherit biases present in the training data \cite{pessach2022review}. Accordingly, extensive efforts have been devoted to improving fairness in machine learning \cite{kamiran2012data, feldman2015certifying, wadsworth2018achieving, scutari2022achieving, zhang2018mitigating, zafar2019fairness, hardt2016equality, corbett2017algorithmic}.

Depending on the stage of model development at which debiasing is introduced, fairness-aware machine learning methods are generally categorized into three groups: pre-processing, in-processing, and post-processing techniques. Pre-processing techniques modify the training data before model fitting to reduce existing biases. Common strategies include adjusting label values \cite{kamiran2009classifying, zhang2017achieving, kamiran2012data}, transforming covariate values \cite{feldman2015certifying}, reweighting training instances \cite{calders2009building}, and resampling the training data \cite{iosifidis2019fae}. In-processing techniques, in contrast, incorporate fairness considerations directly into model training. This can be achieved through adding fairness regularization terms to the objective function \cite{wadsworth2018achieving, scutari2022achieving, zhang2018mitigating} or imposing fairness constraints when optimizing model parameters for predictive performance \cite{zafar2019fairness, celis2019classification}. Post-processing techniques instead modify model predictions after training to satisfy specified fairness requirements \cite{hardt2016equality, corbett2017algorithmic, meursault2025one}. In our work, fairness is addressed during model training, placing the proposed framework in the in-processing category.

Most fair machine learning research has centered on predictive settings such as classification and regression. In our work, we instead address fairness concerns in ITR estimation, an individualized decision-making problem with a causal objective. Treatment decisions depend on potential outcomes under alternative treatments, while only the outcome under the observed treatment is available for each individual. This counterfactual structure distinguishes our setting from conventional fair prediction problems.

\section{Problem Formulation}
We first introduce the preliminaries of the direct learning paradigm for ITR estimation and define the fairness criterion in the multicategory treatment setting. Then, we formulate the corresponding problem of estimating fair ITRs.

\subsection{Preliminaries} \label{subsec:preliminary}
Let $\mathcal{D} = \{(\mathbf{x}_i, s_i, a_i, r_i)\}_{i=1}^n$ denote a dataset of $n$ individuals collected from a randomized trial or an observational study. For each individual, $\mathbf{x}_i \in \mathcal{X} \subseteq \mathbb{R}^p$ is a $p$-dimensional vector of non-sensitive covariates, $s_i \in \mathcal{S}$ is the sensitive attribute, $a_i \in \{1,\dots,K\}$ represents the assigned treatment among $K$ possible options, and $r_i \in \mathbb{R}$ is the observed treatment outcome, also referred to as the reward. The data are assumed to be from a joint distribution $\mathbb{P}_{\mathrm{data}}(\mathbf{X}, S, A, R)$. We consider $K \geq 2$ to allow for multicategory treatments and assume that the reward $R$ is bounded, with larger values indicating more desirable outcomes. Following common practice in the fair machine learning literature \cite{zafar2019fairness, zhang2017achieving, hardt2016equality, xu2019achieving}, we assume a binary sensitive attribute by setting $\mathcal{S} = \{0,1\}$ for ease of discussion, although our framework can be straightforwardly extended to handle multi-class sensitive attributes.

An ITR is a measurable function $d: \mathcal{X} \times \mathcal{S} \rightarrow \{1,\dots,K\}$ that maps individual characteristics $(\mathbf{X}, S)$ to a treatment option. Under the Neyman–Rubin potential outcomes framework \cite{rubin1974estimating}, let $R(k)$ denote the potential outcome that would be observed if an individual were assigned treatment $k \in \{1,\dots,K\}$. For a given ITR $d$, the corresponding potential outcome is $R(d) := \sum_{k=1}^K R(k) I(d(\mathbf{X}, S) = k)$, where $I(\cdot)$ is the indicator function. The value function of $d$ is then defined as
\begin{equation}
    V(d) = \mathrm{E}[R(d)],
\end{equation}
where the expectation is taken over the population distribution of individual characteristics and potential outcomes. However, the value $V(d)$ cannot be directly evaluated from observed data because, for each individual, only the potential outcome corresponding to the treatment actually received is observed. This is the fundamental problem of causal inference \cite{holland1986statistics}. Therefore, identifying $V(d)$ from the observed data relies on the following three common assumptions \cite{rubin1974estimating}:
\begin{enumerate}[label=(\roman*)]
    \item \textbf{Consistency}: If $A = k$, then $R = R(k)$, meaning that the observed outcome equals the potential outcome under the treatment actually received.
    \item \textbf{Unconfoundedness}: $\{R(1),\dots,R(K)\} \independent A \mid \mathbf{X}, S$, that is, conditional on the observed individual characteristics, treatment is independent of all potential outcomes.
    \item \textbf{Positivity}: $0 < \pi_k(\mathbf{x}, s) < 1$, $\forall k \in \{1,\dots,K\}$ and  $\forall (\mathbf{x}, s) \in \mathcal{X} \times \mathcal{S}$, where $\pi_k(\mathbf{x}, s) = \Pr(A = k \mid \mathbf{X}=\mathbf{x}, S=s)$ denotes the propensity score for treatment $k$.
\end{enumerate}
Under these assumptions, the value function $V(d)$ can be identified from $\mathbb{P}_{\mathrm{data}}$ as follows \cite{zhao2012estimating, qian2011performance, huang2019multicategory}:
\begin{equation} \label{eqn:valuefunction}
    V(d) = \mathrm{E}_{(\mathbf{X}, S, A, R) \sim \mathbb{P}_{\mathrm{data}}} \left[ \frac{I(A = d(\mathbf{X}, S))}{\pi_A(\mathbf{X}, S)} R \right],
\end{equation}
where the propensity scores $\pi_A(\mathbf{X}, S)$ are known by design in randomized trials but generally need to be estimated for observational studies. Consequently, an optimal ITR $d^*$ maximizes $V(d)$ over a prespecified class of candidate rules $D_{\mathrm{ITR}}$:
\begin{equation}
    d^* \in \argmax_{d \in D_{\mathrm{ITR}}} V(d).
\end{equation}
For example, $D_{\mathrm{ITR}}$ may consist of all rules parameterized by a given neural network architecture. However, optimizing treatment outcomes alone may result in unfair treatment decisions against protected groups, which motivates incorporating fairness considerations into ITR estimation. Next, we formally define the corresponding fairness criterion.

\subsection{Fairness Criterion for Multicategory ITRs} \label{subsec:fairnesscriterion}
Fairness in algorithmic decision-making can be defined in different ways depending on the application context and the normative considerations \cite{hellman2020measuring}, leading to a broad range of fairness criteria \cite{pessach2022review}. In this work, we consider the principle of equal treatment, according to which access to social and economic opportunities should not depend on sensitive attributes (e.g., the Equal Credit Opportunity Act prohibits discrimination in credit access based on characteristics such as sex and race). Applied to multicategory ITRs, this principle requires subgroups defined by sensitive attributes to have equal probabilities of receiving each treatment option. For the binary sensitive attribute $S \in \{0,1\}$, it can be expressed as:
\begin{equation} 
\Pr(d(\mathbf{X}, S) = k \mid S=1) = \Pr(d(\mathbf{X}, S) = k \mid S=0), 
\end{equation} 
for all $k \in \{1,\dots,K\}$. We formalize this fairness notion below and introduce a quantitative measure of its violation.

\begin{definition}[Fairness of Multicategory ITRs] \label{def:fairness}
Under the population distribution of $(\mathbf{X}, S)$, a multicategory ITR $d: \mathcal{X} \times \mathcal{S} \rightarrow \{1,\dots,K\}$ is fair with respect to the sensitive attribute $S$ if its treatment decision $d(\mathbf{X}, S)$ is independent of $S$, i.e.,
\begin{equation}
    d(\mathbf{X}, S) \independent S.
\end{equation}
In practice, deviations from this fairness criterion are quantified by the treatment-assignment disparity
\begin{equation} \label{eqn:totalvariationdistance}
\Delta_{\mathrm{fair}}(d) = \frac{1}{2} \sum_{k=1}^{K} | \Pr(d(\mathbf{X}, S) = k \mid S=1) - \Pr(d(\mathbf{X}, S) = k \mid S=0) |.
\end{equation}
A smaller value of $\Delta_{\mathrm{fair}}(d)$ indicates greater fairness, with $\Delta_{\mathrm{fair}}(d) = 0$ if and only if $d(\mathbf{X}, S) \independent S$.
\end{definition}

Note that $\Delta_{\mathrm{fair}}(d)$ represents the total variation distance between the conditional distributions of $d(\mathbf{X}, S)$ given $S=1$ and $S=0$. Its value lies in $[0,1]$ regardless of $K$, providing a unified measure of unfairness for multicategory ITRs. This fairness criterion is closely related to demographic parity in fair classification \cite{calders2010three}, while being tailored to the ITR setting. It also generalizes the action fairness notion of \citet{frauen2024fair} from binary to multicategory treatments.

\subsection{Fair ITR Estimation Problem} \label{subsec:problemdefinition}
With the value function and the fairness criterion for multicategory ITRs defined in previous sections, fair ITR estimation should account for both treatment values and fairness in treatment decisions. Next, we formally formulate this estimation problem.

\begin{definition}[Fair ITR Estimation Problem] \label{def:fairITRestimation}
Given the observed data distribution $\mathbb{P}_{\mathrm{data}}$ of $(\mathbf{X}, S, A, R)$, the fair ITR estimation problem aims to find an optimal rule $d(\mathbf{X}, S)$ that maximizes the treatment value subject to the fairness criterion in Definition~\ref{def:fairness}:
\begin{equation} \label{eqn:fairITRestimationconstrainedproblem}
\begin{aligned}
    \max_{d \in D_{\mathrm{ITR}}} \quad& V(d) \\
    \mathrm{s.t.} \quad& \Delta_{\mathrm{fair}}(d) = 0,
\end{aligned}    
\end{equation}
where $V(d)$ is defined in \eqref{eqn:valuefunction} and $\Delta_{\mathrm{fair}}(d)$ is defined in \eqref{eqn:totalvariationdistance}.
\end{definition}

In practice, the exact fairness constraint $\Delta_{\mathrm{fair}}(d) = 0$ in \eqref{eqn:fairITRestimationconstrainedproblem} may be overly restrictive. A relaxed constraint $\Delta_{\mathrm{fair}}(d) \leq \tau$ can instead be considered, where $\tau \in (0,1]$ represents the maximum tolerated disparity in treatment assignments across groups and is chosen according to the application-specific fairness requirement. Moreover, the maximization of the value function $V(d)$ can be equivalently reformulated as the following minimization problem:
\begin{equation} \label{eqn:minimizeweightederror}
    \min_{d \in D_{\mathrm{ITR}}} \quad \mathrm{E}_{(\mathbf{X}, S, A, R) \sim \mathbb{P}_{\mathrm{data}}}
    \left[ \frac{I(A \neq d(\mathbf{X}, S))}{\pi_A(\mathbf{X}, S)} R \right],
\end{equation}
since the two objectives sum to a quantity that is independent of $d$. This equivalent formulation is widely used by direct ITR learning methods \cite{zhao2012estimating, huang2019multicategory, zhang2020multicategory}. It closely resembles a weighted classification problem and can therefore be approached within the general paradigm of multicategory classification. Unless otherwise stated, the remainder of this paper adopts the minimization form in \eqref{eqn:minimizeweightederror} when optimizing the treatment value.

\section{Methodology}
\begin{figure*}[htbp]
  \centering
  \includegraphics[width=0.83\textwidth]{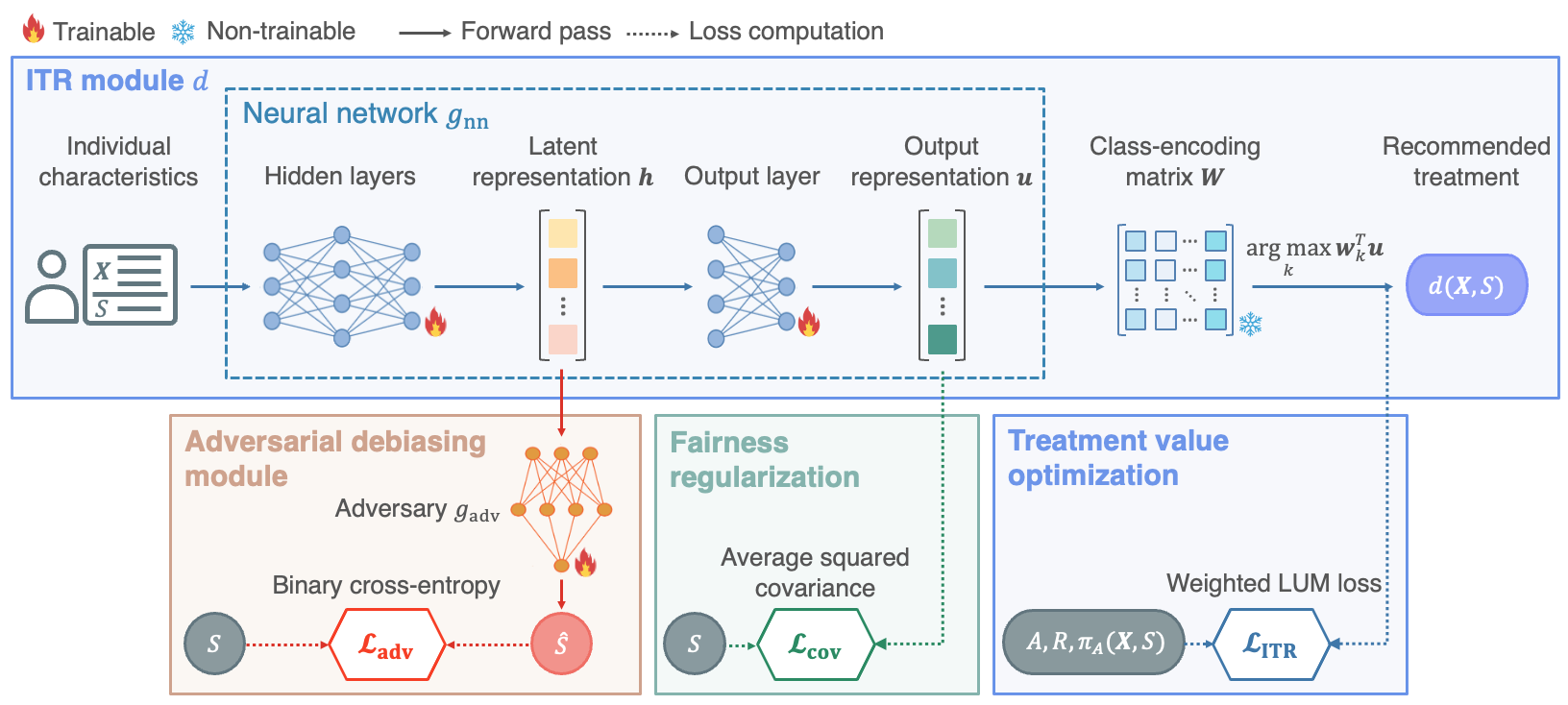}
  \caption{Overall architecture of the proposed FairITR framework.}
  \Description{Architecture of FairITR showing the neural network-based ITR module, adversarial debiasing module, and covariance-based fairness regularization used to jointly learn treatment decisions and improve fairness with respect to the sensitive attribute.}
  \label{fig:flowchart}
\end{figure*}

In this section, we introduce a fairness-aware learning framework for multicategory ITRs, \textbf{FairITR}, to solve the problem in \eqref{eqn:fairITRestimationconstrainedproblem} using neural network-based treatment rules. The overall design of FairITR is shown in Figure~\ref{fig:flowchart} and consists of three main components: an ITR module $d$ for decision-making, an adversarial debiasing module $g_{\mathrm{adv}}$ for mitigating unfairness associated with the sensitive attribute, and a fairness regularization term for stabilizing the debiasing process.

The ITR module $d$ is built on a neural network $g_{\mathrm{nn}}$ that maps individual characteristics $(\mathbf{X}, S)$ to an output representation. This representation is then transformed into a treatment score vector through a predefined, non-trainable class-encoding matrix, and the treatment with the highest score is selected as the recommended option. To promote fair treatment decisions, the adversarial module $g_{\mathrm{adv}}$ seeks to predict the sensitive attribute $S$ using the latent representation from the last hidden layer of $g_{\mathrm{nn}}$, while $g_{\mathrm{nn}}$ is trained adversarially to make such prediction difficult. This minimax game encourages the latent representation to contain less sensitive information, thereby reducing its influence on the resulting treatment decisions. We further provide theoretical results characterizing the fairness properties induced by this adversarial learning mechanism. To better stabilize fairness performance, we incorporate a fairness regularization term into the training objective based on the covariance between $S$ and the output representation from $g_{\mathrm{nn}}$. This term provides additional support for improving fairness, particularly when adversarial training exhibits instability. For the remainder of this section, we describe each component of FairITR in detail.

\subsection{ITR Module} \label{sec:ITRModule}
Based on the formulation in Section~\ref{subsec:problemdefinition}, the treatment value of an ITR $d$ is optimized through the classification-type objective in \eqref{eqn:minimizeweightederror}. However, the indicator function in this objective is discontinuous, making direct optimization computationally difficult. A suitable surrogate loss function is therefore typically used in practice. To address this problem in the multicategory setting, we build the ITR module on the angle-based classification framework proposed by \citet{zhang2014multicategory}. This framework has been successfully adapted to multicategory ITR estimation and provides a computationally efficient approach for handling multiple treatment options \cite{zhang2020multicategory, qi2020multi}.

In particular, for $K \geq 2$ treatment options, the ITR module employs a neural network $g_{\mathrm{nn}}: \mathcal{X} \times \mathcal{S} \rightarrow \mathbb{R}^{K-1}$ that maps individual characteristics $(\mathbf{X}, S)$ to a $(K-1)$-dimensional representation vector $\mathbf{u} = g_{\mathrm{nn}}(\mathbf{X}, S) = (u_1, \dots, u_{K-1})^T$. Under the angle-based classification framework, we represent each treatment option $k \in \{1,\dots,K\}$ by a $(K-1)$-dimensional class-encoding vector $\mathbf{w}_k \in \mathbb{R}^{K-1}$. These vectors are defined as
\begin{equation}
    \mathbf{w}_k =
    \begin{cases}
        \frac{1}{\sqrt{K-1}} \mathbf{1}_{K-1} & k=1, \\
        -\frac{(1+\sqrt{K})}{\sqrt{(K-1)^3}} \mathbf{1}_{K-1} + \sqrt{\frac{K}{K-1}} \mathbf{e}_{k-1} & 2 \leq k \leq K,
    \end{cases}
\end{equation}
where $\mathbf{1}_{K-1}$ denotes the $(K-1)$-dimensional vector of ones, and $\mathbf{e}_{k-1}$ denotes the $(k-1)$th standard basis vector in $\mathbb{R}^{K-1}$. Let $\mathbf{W} = (\mathbf{w}_1,\dots,\mathbf{w}_K) \in \mathbb{R}^{(K-1) \times K}$ represent the treatment class-encoding matrix. The vectors $\{\mathbf{w}_k\}_{k=1}^K$ form the vertices of a regular simplex centered at the origin in $\mathbb{R}^{K-1}$. They satisfy $\|\mathbf{w}_k\|_2=1$ and $\mathbf{w}_k^T\mathbf{w}_{k'} = - \frac{1}{K-1}$ for $k \neq k'$. Accordingly, the treatment score vector is defined as $\mathbf{z} = \mathbf{W}^T \mathbf{u} \in \mathbb{R}^K$, in which the $k$th element represents the score for treatment option $k$. The recommended treatment is then selected as the option with the largest score. Hence, the ITR $d$ takes the form
\begin{equation}
    d(\mathbf{X}, S) = \argmax_{k \in \{1,\dots,K\}} \mathbf{w}_k^T g_{\mathrm{nn}}(\mathbf{X}, S).
\end{equation}

To address the discontinuity of the indicator function in \eqref{eqn:minimizeweightederror}, we adopt the large-margin unified machine (LUM) loss introduced in \citet{liu2011hard} as the surrogate loss. The LUM loss involves two hyperparameters, providing flexibility across a broad class of margin-based surrogate losses. It is defined as
\begin{equation} \label{eqn:lumloss}
    l(t) =
    \begin{cases}
        1 - t & t < \frac{\eta}{1 + \eta}, \\
        \frac{1}{1 + \eta} \left( \frac{\zeta}{(1 + \eta)t - \eta + \zeta} \right)^\zeta & t \geq \frac{\eta}{1 + \eta},
    \end{cases}
\end{equation}
where $\zeta > 0$ and $\eta \geq 0$ are hyperparameters that determine the specific loss function within the LUM family. For example, as $\eta \rightarrow \infty$, the LUM loss approaches the hinge loss used in support vector machines, while setting $\zeta = 1$ and $\eta = 1$ yields the distance-weighted discrimination loss of \citet{marron2007distance}.

One remaining challenge in approaching the objective in \eqref{eqn:minimizeweightederror} as a weighted classification problem is that the observed reward $r_i$ can be negative, resulting in negative sample weights. Intuitively, when an individual receives a large positive reward under the observed treatment $a_i$, the corresponding treatment score $\mathbf{w}_{a_i}^T g_{\mathrm{nn}}(\mathbf{x}_i, s_i)$ should be encouraged to increase. Conversely, when the observed reward is negative, assigning the same treatment is undesirable, and the corresponding score should be encouraged to decrease. Following the strategy adopted in \citet{zhang2020multicategory} for handling negative rewards, we incorporate the sign of $r_i$ into the classification margin while using $|r_i|$ as a nonnegative sample weight. The resulting empirical loss for training $g_{\mathrm{nn}}$ is
\begin{equation} \label{eqn:lossitr}
    \mathcal{L}_{\mathrm{ITR}} = \frac{1}{n} \sum_{i=1}^n \frac{|r_i|}{\pi_{a_i}(\mathbf{x}_i, s_i)} l\left( \mathrm{sign}(r_i) \mathbf{w}_{a_i}^T g_{\mathrm{nn}}(\mathbf{x}_i, s_i) \right),
\end{equation}
where $l(\cdot)$ is the LUM loss in \eqref{eqn:lumloss}, and $\mathrm{sign}(\cdot)$ denotes the sign function. For observations with negative rewards, the sign adjustment reverses the classification margin, thereby favoring lower scores for the observed treatment.

\subsection{Adversarial Debiasing Module}
Without explicit fairness interventions, the output representations $\mathbf{u}$ learned by $g_{\mathrm{nn}}$ may encode biased information with respect to the sensitive attribute, leading the ITR $d$ to produce unfair treatment decisions across protected groups. Such unfairness often originates from biases present in the training data, which can be readily captured by $g_{\mathrm{nn}}$ given the strong expressive power of neural networks. Recently, adversarial learning has emerged as an effective approach for mitigating biases by guiding neural networks to learn fair representations that contain less sensitive information \cite{wadsworth2018achieving, xu2019achieving, zhang2018mitigating}. In FairITR, we introduce an adversarial debiasing module to address the fairness constraint in problem \eqref{eqn:fairITRestimationconstrainedproblem}.

We consider the model $g_{\mathrm{nn}}$ as a composition of two components. Let $g_{\mathrm{hid}}: \mathcal{X} \times \mathcal{S} \rightarrow \mathbb{R}^q$ denote the mapping from the input to the last hidden layer, and let $g_{\mathrm{out}}: \mathbb{R}^q \rightarrow \mathbb{R}^{K-1}$ denote the output layer. The $q$-dimensional latent representation is given by $\mathbf{h} = g_{\mathrm{hid}}(\mathbf{X}, S)$. It is subsequently transformed by $g_{\mathrm{out}}$ into the output representation $\mathbf{u} = g_{\mathrm{out}}(\mathbf{h}) = g_{\mathrm{nn}}(\mathbf{X}, S)$. Equivalently, $g_{\mathrm{nn}} = g_{\mathrm{out}} \circ g_{\mathrm{hid}}$. Let $\boldsymbol{\theta}_{\mathrm{hid}}$ and $\boldsymbol{\theta}_{\mathrm{out}}$ denote the learnable parameters of $g_{\mathrm{hid}}$ and $g_{\mathrm{out}}$, respectively, with the parameters of $g_{\mathrm{nn}}$ collectively denoted by $\boldsymbol{\theta}_{\mathrm{nn}}=\{\boldsymbol{\theta}_{\mathrm{hid}}, \boldsymbol{\theta}_{\mathrm{out}}\}$.

In the adversarial debiasing module, an adversary $g_{\mathrm{adv}}: \mathbb{R}^q \rightarrow [0,1]$ takes the latent representation $\mathbf{h}$ as input and outputs $\hat{S} = g_{\mathrm{adv}}(\mathbf{h})$, where $\hat{S}$ is an estimate of $\Pr(S=1 \mid \mathbf{h})$. Here $g_{\mathrm{adv}}$ aims to predict the sensitive attribute $S$ accurately, whereas $g_{\mathrm{hid}}$ is trained adversarially to make this prediction difficult by reducing the sensitive information encoded in $\mathbf{h}$. Meanwhile, $g_{\mathrm{hid}}$ must preserve information in $\mathbf{h}$ that is useful for decision-making. Such an adversarial training process defines a minimax game between $g_{\mathrm{hid}}$ and $g_{\mathrm{adv}}$ as
\begin{equation} \label{eqn:minimaxgame}
\begin{aligned}
    \min_{\boldsymbol{\theta}_{\mathrm{hid}}} \max_{\boldsymbol{\theta}_{\mathrm{adv}}} \big\{ \Psi(g_{\mathrm{hid}}, g_{\mathrm{adv}}) :=&\; \mathrm{E}_{\mathbf{h} \sim f(\mathbf{h} \mid S=1)} [\log(g_{\mathrm{adv}}(\mathbf{h}))] \\
    &+ \mathrm{E}_{\mathbf{h} \sim f(\mathbf{h} \mid S=0)} [\log(1-g_{\mathrm{adv}}(\mathbf{h}))] \big\},
\end{aligned}
\end{equation}
where $\Psi(g_{\mathrm{hid}}, g_{\mathrm{adv}})$ denotes the objective of the two-player minimax game, $f(\mathbf{h} \mid S=1)$ and $f(\mathbf{h} \mid S=0)$ represent the conditional distributions of the latent representation given the sensitive attribute, and $\boldsymbol{\theta}_{\mathrm{adv}}$ denotes the learnable parameters of $g_{\mathrm{adv}}$.

We apply adversarial debiasing to $\mathbf{h} \in \mathbb{R}^q$ rather than directly to the output representation $\mathbf{u} \in \mathbb{R}^{K-1}$. This is because the latent space is typically higher-dimensional than the output space, particularly when $K$ is small. Operating in this richer representation space can provide greater flexibility to remove sensitive information while preserving information important for treatment decision-making. In the following proposition, we establish a connection between the global optimum of the game in \eqref{eqn:minimaxgame} and the fairness criterion in Definition~\ref{def:fairness}, providing theoretical justification for how adversarial debiasing on $\mathbf{h}$ improves the fairness of an ITR.

\begin{proposition} \label{thm:advlearning}
The global optimum of the minimax game \eqref{eqn:minimaxgame} is characterized by $f(\mathbf{h} \mid S=1) = f(\mathbf{h} \mid S=0)$ almost everywhere. At this optimum, for any output layer $g_{\mathrm{out}}$, the resulting ITR
\begin{equation*}
    d(\mathbf{X}, S) = \argmax_{k \in \{1,\dots,K\}} \mathbf{w}_k^T g_{\mathrm{out}}(\mathbf{h})
\end{equation*}
satisfies $d(\mathbf{X}, S) \independent S$, and is therefore fair according to Definition~\ref{def:fairness}.
\end{proposition}

We defer the proof to Appendix~\ref{appendixsec:proof}. The proposition shows that, at the global optimum of the adversarial debiasing optimization, the resulting ITR produces treatment decisions that are independent of the sensitive attribute, thereby preventing discrimination across protected groups. Based on $\Psi(g_{\mathrm{hid}}, g_{\mathrm{adv}})$, the empirical loss function used to train the adversarial debiasing module is
\begin{equation} \label{eqn:lossadv}
    \mathcal{L}_{\mathrm{adv}} = -\frac{1}{n} \sum_{i=1}^n \left[s_i \log(\hat{s}_i) + (1-s_i) \log(1-\hat{s}_i) \right].
\end{equation}

\subsection{Fairness Regularization}
The theoretical fairness properties in Proposition~\ref{thm:advlearning} rely on reaching the global optimum of the adversarial debiasing. In practice, however, adversarial training is known to suffer from instability and convergence difficulties \cite{mescheder2018training, bang2018improved}, which may prevent the learned ITR from fully attaining the ideal fairness condition. To provide an additional safeguard against such instability and stabilize fairness performance, we introduce a fairness regularization term directly on the output representation $\mathbf{u}$ of $g_{\mathrm{nn}}$. Specifically, we consider the covariance between each element of $\mathbf{u}$ and the sensitive attribute $S$. For binary $S \in \{0,1\}$, the covariance for the $j$th output dimension satisfies $\mathrm{Cov}(u_j, S) = \Pr(S=1)\Pr(S=0)[\mathrm{E}(u_j \mid S=1) - \mathrm{E}(u_j \mid S=0)]$. Thus, reducing $|\mathrm{Cov}(u_j, S)|$ narrows the difference in the mean of $u_j$ across the two protected groups. Based on this observation, we construct the fairness regularization term using the average squared covariance across the $K-1$ output dimensions:
\begin{equation}
    \frac{1}{K-1} \sum_{j=1}^{K-1} \mathrm{Cov}(u_j, S)^2.
\end{equation}
In practice, each covariance is estimated by its sample counterpart. The empirical form of the regularization used for training is
\begin{equation} \label{eqn:losscov}
    \mathcal{L}_{\mathrm{cov}} = \frac{1}{K-1} \sum_{j=1}^{K-1} \left[ \frac{1}{n-1} \sum_{i=1}^n (u_{i,j} - \bar{u}_{\cdot,j}) (s_i - \bar{s}) \right]^2,
\end{equation}
where $u_{i,j}$ denotes the $j$th element of the output representation for individual $i$, $\bar{u}_{\cdot,j} = \frac{1}{n} \sum_{i=1}^n u_{i,j}$ is its sample mean, and $\bar{s} = \frac{1}{n} \sum_{i=1}^n s_i$ is the sample mean of the sensitive attribute.

\subsection{FairITR Training Objective}
Having defined the empirical losses for the ITR module, adversarial debiasing module, and covariance-based fairness regularization, we integrate them into the overall training objective of FairITR:
\begin{equation} \label{eqn:overallobjective}
    \min_{\boldsymbol{\theta}_{\mathrm{nn}}} \max_{\boldsymbol{\theta}_{\mathrm{adv}}} \; \mathcal{L}_{\mathrm{ITR}} + \alpha \mathcal{L}_{\mathrm{cov}} - \beta \mathcal{L}_{\mathrm{adv}},
\end{equation}
where $\alpha$ and $\beta$ are two nonnegative hyperparameters that control the strengths of fairness regularization and adversarial debiasing, respectively. We optimize this objective through alternating updates of $\boldsymbol{\theta}_{\mathrm{nn}}$ and $\boldsymbol{\theta}_{\mathrm{adv}}$. At each iteration, we first fix $\boldsymbol{\theta}_{\mathrm{adv}}$ and update $\boldsymbol{\theta}_{\mathrm{nn}}$ to minimize the overall objective, which simultaneously reduces $\mathcal{L}_{\mathrm{ITR}}$ and $\mathcal{L}_{\mathrm{cov}}$ while increasing $\mathcal{L}_{\mathrm{adv}}$ to make the sensitive attribute more difficult to predict. We then fix $\boldsymbol{\theta}_{\mathrm{nn}}$ and update $\boldsymbol{\theta}_{\mathrm{adv}}$ by maximizing $(- \beta \mathcal{L}_{\mathrm{adv}})$, equivalently minimizing $\mathcal{L}_{\mathrm{adv}}$, such that the adversary improves its prediction of the sensitive attribute. In implementation, we use two separate Adam optimizers \cite{kingma2014adam} for $g_{\mathrm{nn}}$ and $g_{\mathrm{adv}}$.

\section{Experiments}
In this section, we conduct numerical experiments on real-world datasets to evaluate the effectiveness of FairITR in achieving favorable treatment outcomes while mitigating unfairness in treatment decisions. Specifically, we answer the following research questions:
\begin{itemize}[topsep=4pt]
    \item \textbf{RQ1}: How effective is the proposed FairITR in learning fair multicategory ITRs compared with existing baselines?
    \item \textbf{RQ2}: How do the adversarial debiasing and the fairness regularization contribute to the performance of FairITR?
    \item \textbf{RQ3}: How sensitive is the performance of FairITR to the hyperparameters in the overall training objective \eqref{eqn:overallobjective}?
\end{itemize}

\subsection{Experimental Setup}
\subsubsection{Datasets}
To evaluate the practical performance of FairITR comprehensively, we use three publicly available datasets covering both randomized trials and observational studies: RHC, Email, and IST. Table~\ref{tab:datasetsummary} provides a brief summary of these datasets, and their details are described as follows.

\begin{table}[tbp]
\small
\caption{Summary of the datasets used in the experiments.}
\label{tab:datasetsummary}
\begin{tabular}{lccc}
    \toprule
    \textbf{Dataset} & \textbf{RHC} & \textbf{Email} & \textbf{IST} \\
    \midrule
    \textbf{Sample Size} $n$ & 5,735 & 64,000 & 19,435 \\
    \textbf{\# Treatments} $K$ & 2 & 3 & 4 \\
    \textbf{\# Covariates} $p$ & 58 & 7 & 19 \\
    \textbf{Study Design} & Observational & Randomized & Randomized \\
    \bottomrule
\end{tabular}
\end{table}

\textbf{RHC} \cite{lee2021atbounds}. This dataset is an observational study of right heart catheterization (RHC), a diagnostic procedure used to directly assess cardiac function in critically ill patients. The treatment is binary ($K=2$), indicating whether RHC is applied within 24 hours of admission. The reward is 30-day survival after admission, coded as 1 for survival and 0 otherwise. We use medical insurance type (private vs. non-private) as the binary sensitive attribute. For the non-sensitive covariates $\mathbf{X}$, we remove variables with missing values and binary variables with extremely low prevalence. This pre-processing step retains 58 of the original 66 variables in $\mathbf{X}$.

\textbf{Email} \cite{hillstrom2008email}. This dataset is from a randomized email marketing campaign, and customers were randomly assigned to one of three marketing strategies, resulting in a multicategory treatment setting ($K=3$). Following \citet{lee2024effective}, the reward is whether a customer visits the website within two weeks after receiving the campaign email (1 = Yes and 0 = No). We use whether the customer purchased women's merchandise in the previous year as the sensitive attribute $S$. The remaining seven historical customer features are used as $\mathbf{X}$.

\textbf{IST} \cite{sandercock2011international}. This dataset comes from the International Stroke Trial (IST), a multi-center randomized trial investigating the effects of aspirin and heparin in acute stroke. The combinations of aspirin and heparin use lead to four treatment options ($K=4$). The reward is a binary indicator of being free from death or dependency at six months (1 = Yes and 0 = No). Following prior studies on this dataset \cite{bouvier2024machine, nguyen2020counterfactual}, we use the same 19 baseline covariates, except that we redefine the region variable as a binary indicator of whether a patient's country ranks among the top 30 countries by IMF GDP per capita \cite{IMF2026GDP} and treat it as $S$. Since these studies focus on aspirin only, we include heparin treatment history as an additional covariate to accommodate our joint consideration of aspirin and heparin.

\subsubsection{Baselines} \label{subsubsec:baselines}
We compare the proposed FairITR with several representative and state-of-the-art ITR learning methods, including both conventional approaches without explicit fairness considerations and fairness-aware methods. The baselines include:
\begin{itemize}[leftmargin=*]
    \item \textbf{$\boldsymbol{l_1}$-PLS} \cite{qian2011performance}: A Q-learning method that estimates $\mathrm{E}(R \mid \mathbf{X}, S, A)$ and selects the treatment with the highest estimated outcome. It uses the LASSO regularization for variable selection when fitting the outcome model.
    \item \textbf{OWL} \cite{zhao2012estimating}: A direct method that formulates ITR estimation as a weighted classification problem. It applies the hinge loss as a surrogate for the classification error and solves the optimization problem using support vector machine techniques.
    \item \textbf{OWL-MD} \cite{huang2019multicategory}: This method is an extension of OWL from binary to multicategory treatments. It employs the multinomial deviance loss as a surrogate for the multicategory classification error.
    \item \textbf{OML} \cite{zhang2020multicategory}: OML generalizes the hinge loss adopted by OWL to a broader class of margin-based surrogate losses. It also introduces an inverted loss strategy to handle negative rewards.
    \item \textbf{MOML} \cite{zhang2020multicategory}: A generalization of OML to settings with multiple treatments using an angle-based classification framework.
    \item \textbf{UF-ITR} \cite{lee2024effective}: We refer to the unified framework for ITR learning in \citet{lee2024effective} as UF-ITR for brevity. It integrates distributional covariate balancing weights, parameter regularization, variable screening, outcome augmentation, and inverse variance weighting into ITR estimation.
    \item \textbf{FairPol} \cite{frauen2024fair}: This is a fairness-aware method for learning neural network-based ITRs using a two-stage design. It first learns fair representations of individual characteristics and then estimates an ITR from these representations. Such a two-stage design separates fairness mitigation from treatment outcome optimization. The original method models stochastic ITRs, which recommend a probability distribution over treatments, and is limited to binary treatments. For a fair comparison under our multicategory deterministic ITR setting, where an ITR directly assigns a single treatment option, we retain its representation learning stage and build the second stage on the ITR module described in Section~\ref{sec:ITRModule}.
\end{itemize}
Except for FairPol, the remaining baselines do not explicitly address fairness concerns. For each of these methods, we therefore consider an additional fairness-through-unawareness variant that excludes $S$ from the model input, denoted by the suffix ``-FTU'' (e.g., OWL-FTU). Also, since OWL and OML are limited to binary treatments, we use them for the RHC dataset, while their multicategory counterparts, OWL-MD and MOML, are used for the Email and IST datasets.

\subsubsection{Evaluation Metrics}
For evaluating treatment outcomes, following prior ITR literature \cite{huang2019multicategory, zhao2012estimating, zhang2020multicategory}, we use the empirical value function. This metric is denoted by \textbf{Value} and is calculated as
\begin{equation}
    \frac{\sum_{i=1}^n I(a_i = d(\mathbf{x}_i, s_i)) r_i / \pi_{a_i}(\mathbf{x}_i, s_i)}{\sum_{i=1}^n I(a_i = d(\mathbf{x}_i, s_i)) / \pi_{a_i}(\mathbf{x}_i, s_i)}.
\end{equation}
To evaluate fairness, we use the empirical counterpart of $\Delta_{\mathrm{fair}}(d)$ in \eqref{eqn:totalvariationdistance}. Specifically, each conditional probability $\Pr(d(\mathbf{X}, S) = k \mid S=s)$ is estimated by $\frac{1}{n_s} \sum_{i: s_i=s} I(d(\mathbf{x}_i, s_i) = k)$, where $n_s$ represents the sample size of the group $S=s$. Substituting these estimates into $\Delta_{\mathrm{fair}}(d)$ gives its empirical value.

\subsubsection{Implementation Details} \label{subsubsec:implmentationdetails}
We randomly split each dataset into training and testing sets with a 60/40 ratio. Every method is trained on the training set and evaluated on the testing set. We implement FairITR in PyTorch \cite{paszke2017automatic}. The neural network $g_{\mathrm{nn}}$ is implemented as a multilayer perceptron with one hidden layer of dimension 64 for RHC, one hidden layer of dimension 32 for Email, and two hidden layers of dimension 32 for IST. The adversary $g_{\mathrm{adv}}$ is implemented as a linear classifier. As for the LUM loss $l(\cdot)$ appearing in \eqref{eqn:lossitr}, we fix $\zeta=1$ and $\eta=0$ across all experiments, following the empirical results from \citet{liu2011hard}, where this specification was shown to perform well across a range of settings. The hyperparameters $\alpha$ and $\beta$ in the FairITR training objective are determined separately for each dataset through cross-validation with grid search. We run all experiments 10 times and report the mean and standard error of each evaluation metric. Additional implementation details for FairITR and the baseline methods are provided in the Appendix~\ref{appendixsec:experimentdetails}. Our code will be made publicly available upon completion of the review process.

\subsection{Comparative Performance (RQ1)}

\begin{table*}[htbp]
  \small
  \caption{Performance comparison between FairITR and the baselines. Standard errors are shown in parentheses. $\uparrow$ indicates that a higher value is preferred, and $\downarrow$ indicates the opposite. The best and runner-up results are highlighted in \textbf{bold} and \underline{underlined}, respectively. ``---'' indicates that the method is not used for the corresponding dataset because of the treatment setting.}
  \label{tab:performancecomparisonRQ1}
  \begin{tabular}{l|l|cc cc cc}
    \toprule
    \multicolumn{1}{l}{} & \multicolumn{1}{l}{} & \multicolumn{2}{c}{\textbf{RHC}} & \multicolumn{2}{c}{\textbf{Email}} & \multicolumn{2}{c}{\textbf{IST}} \\
    \cmidrule(lr){3-4} \cmidrule(lr){5-6} \cmidrule(lr){7-8}
    \multicolumn{1}{l}{\textbf{Category}} & \multicolumn{1}{l}{\textbf{Method}} & \textbf{Value} ($\uparrow$) & $\boldsymbol{\Delta_{\mathrm{fair}}}$ ($\downarrow$) & \textbf{Value} ($\uparrow$) & $\boldsymbol{\Delta_{\mathrm{fair}}}$ ($\downarrow$) & \textbf{Value} ($\uparrow$) & $\boldsymbol{\Delta_{\mathrm{fair}}}$ ($\downarrow$) \\
    \midrule
    \multirow{6}{*}{\textbf{Conventional}} & $l_1$-PLS & 0.3545 (0.005) & 0.0997 (0.042) & 0.1832 (0.001) & 0.3271 (0.038) & 0.3785 (0.003) & 0.0671 (0.025) \\
    & OWL & 0.3453 (0.005) & 0.0461 (0.009) & --- & --- & --- & --- \\
    & OWL-MD & --- & --- & 0.1833 (0.001) & 0.4413 (0.042) & 0.3730 (0.003) & 0.1868	(0.024)  \\
    & OML & 0.3493 (0.005) & 0.0989 (0.018) & --- & --- & --- & --- \\
    & MOML & --- & --- & \textbf{0.1840 (0.001)} & 0.4984 (0.037) & 0.3760 (0.003) & 0.2240 (0.037) \\
    & UF-ITR & \textbf{0.3618 (0.005)} & 0.0460 (0.011) & 0.1829 (0.004) & 0.1692 (0.085) & 0.3772 (0.002) & 0.1194 (0.021) \\
    \midrule
    \multirow{7}{*}{\textbf{Fairness-Adjusted}} & $l_1$-PLS-FTU & 0.3539 (0.005) & 0.0863 (0.033) & 0.1831 (0.001) & 0.2767 (0.033) & 0.3774 (0.002) & \underline{0.0442 (0.012)} \\
    & OWL-FTU & 0.3459 (0.004) & 0.0715 (0.012) & --- & --- & --- & --- \\
    & OWL-MD-FTU & --- & --- & 0.1831 (0.001) & 0.4187 (0.043) & 0.3753 (0.004) & 0.0515	(0.007) \\
    & OML-FTU & 0.3484 (0.005) & 0.1075 (0.017) & --- & --- & --- & --- \\
    & MOML-FTU & --- & --- & 0.1834	(0.001) & 0.4784 (0.033) & 0.3732 (0.003) & 0.0760 (0.009) \\
    & UF-ITR-FTU & \underline{0.3610 (0.006)} & 0.0447 (0.015) & \underline{0.1838 (0.004)} & 0.2123 (0.101) & 0.3771 (0.002) & 0.0661 (0.011) \\
    & FairPol & 0.3522 (0.003) & \underline{0.0259 (0.012)} & 0.1825 (0.001) & \textbf{0.0314 (0.016)} & \underline{0.3819 (0.003)} & 0.0582 (0.014) \\
    \midrule
    \textbf{Ours} & FairITR & 0.3569 (0.003) & \textbf{0.0116 (0.003)} & 0.1831 (0.001) & \underline{0.0475 (0.011)} & \textbf{0.3837 (0.002)} & \textbf{0.0370 (0.005)} \\
    \bottomrule
  \end{tabular}
\end{table*}

To answer RQ1, we estimate ITRs on each dataset using FairITR and the aforementioned baseline methods, and compare their treatment outcome value and decision fairness. The results are summarized in Table~\ref{tab:performancecomparisonRQ1}. From the table, we make the following four observations.

First, conventional ITR methods that do not explicitly consider fairness can achieve favorable treatment values but often exhibit substantial fairness disparities. For example, MOML achieves the highest Value on Email, while its $\Delta_{\mathrm{fair}}$ is close to 0.5, highlighting the need to account for fairness in ITR estimation.

Second, the FTU strategy does not consistently improve fairness for conventional methods. For example, OWL-FTU and OML-FTU lead to higher $\Delta_{\mathrm{fair}}$ than their original counterparts on RHC, and several FTU variants provide only limited improvements on Email. In contrast, FairPol and FairITR explicitly incorporate fairness into ITR learning and generally achieve substantially lower $\Delta_{\mathrm{fair}}$ values.

Third, FairITR demonstrates strong overall performance in terms of both fairness and treatment outcomes. It achieves the lowest $\Delta_{\mathrm{fair}}$ on RHC and IST and the second-lowest on Email. At the same time, these substantial fairness improvements come with little apparent sacrifice in treatment value: FairITR is exceeded only by UF-ITR and UF-ITR-FTU on RHC, performs comparably to most baselines on Email, and achieves the highest Value on IST.

Fourth, FairITR overall provides a more favorable balance between treatment outcome value and fairness than FairPol, particularly on RHC and IST, where it attains both higher Value and lower $\Delta_{\mathrm{fair}}$. This result suggests an advantage of jointly optimizing treatment outcomes and fairness within a unified learning framework.

Overall, these results demonstrate the effectiveness of FairITR in improving fairness in ITR estimation while maintaining competitive treatment values relative to existing baselines.

\subsection{Ablation Study (RQ2)}
\begin{figure}[htbp]
  \centering
  \begin{subfigure}[t]{0.495\columnwidth}
    \centering
    \includegraphics[width=\linewidth]{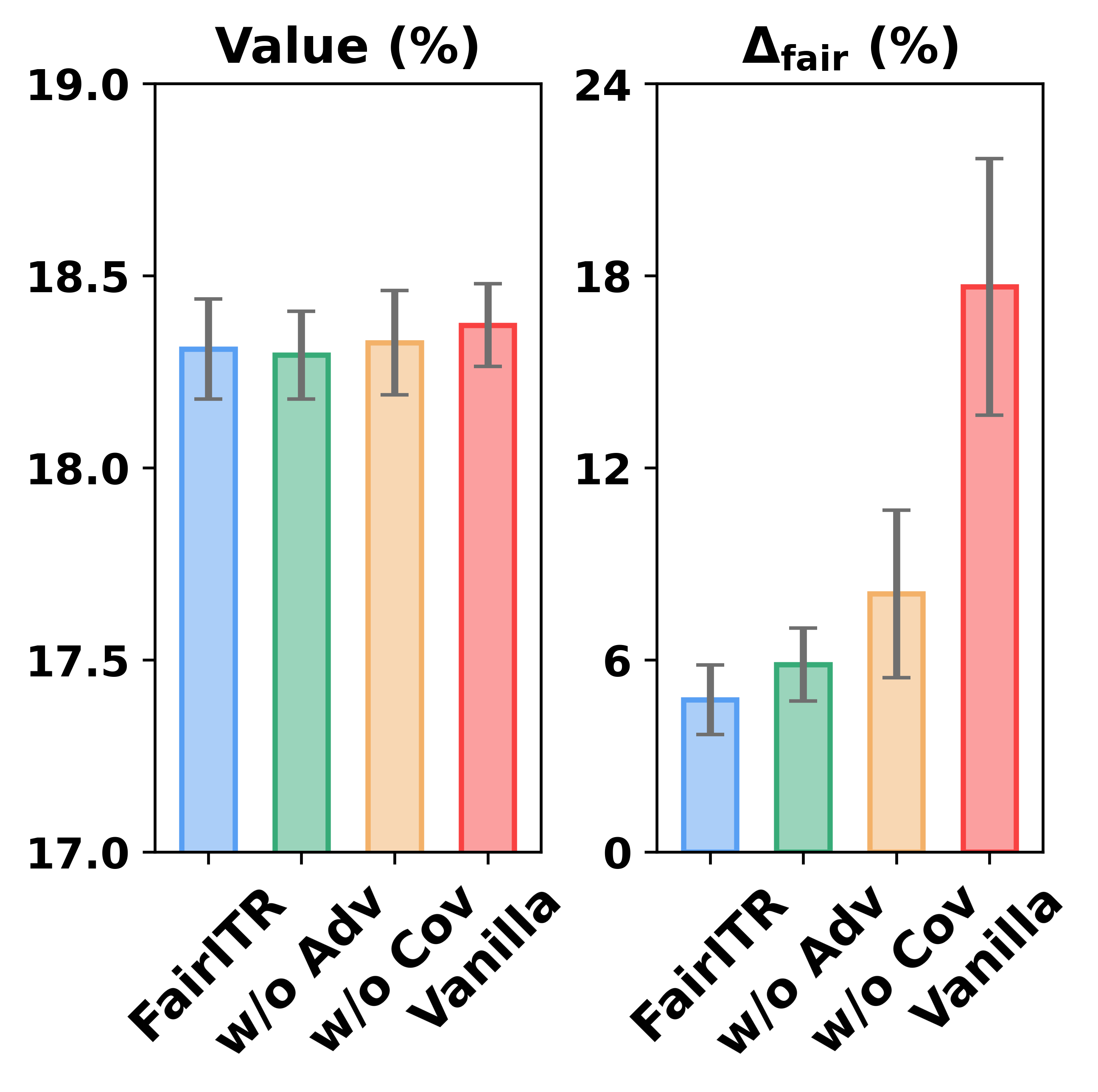}
    \caption{Email}
    \label{fig:ablation_email}
  \end{subfigure}
  \hfill 
  \begin{subfigure}[t]{0.495\columnwidth}
    \centering
    \includegraphics[width=\linewidth]{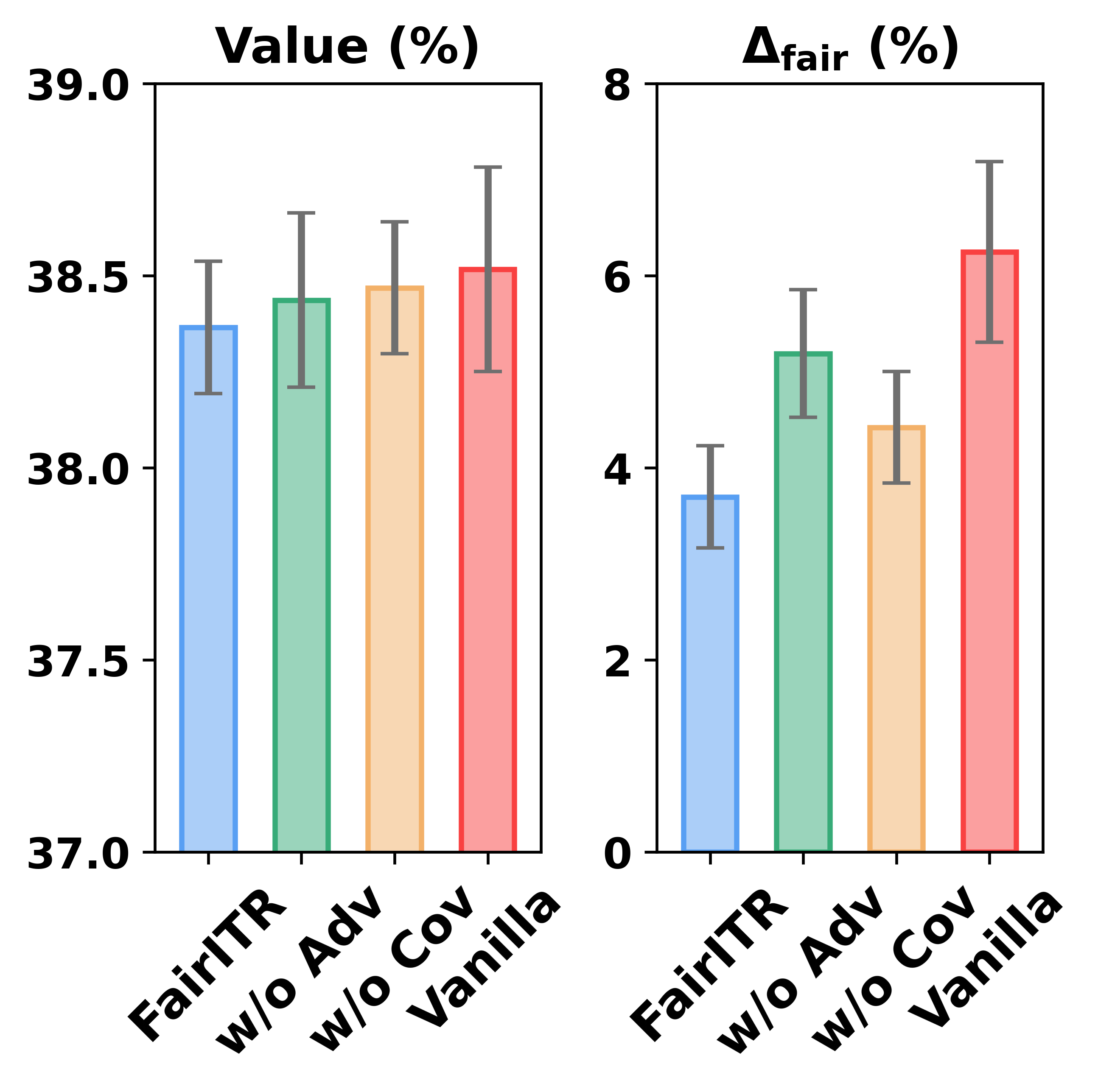}
    \caption{IST}
    \label{fig:ablation_ist}
  \end{subfigure}
\caption{Ablation study of the adversarial debiasing module and covariance-based fairness regularization term in FairITR. All values are shown on a percentage scale.}
\Description{Bar charts comparing FairITR and three ablation variants on the Email and IST datasets in terms of treatment value and fairness disparity.}
\label{fig:ablationstudy}
\end{figure}

The adversarial debiasing and fairness regularization components in FairITR are both designed to improve the fairness of treatment decisions. To assess their contributions to the performance of FairITR, we conduct a set of ablation experiments. For this analysis, we construct three incomplete variants of FairITR: 1) w/o Adv (without adversarial debiasing), 2) w/o Cov (without covariance-based fairness regularization), and 3) Vanilla (without either component). Except for the removed component(s), the three variants follow the same implementation settings as FairITR described in Section~\ref{subsubsec:implmentationdetails}. For w/o Adv and w/o Cov, the remaining hyperparameter ($\alpha$ or $\beta$) is selected via cross-validation with grid search. As illustrative examples, we evaluate these variants on the multicategory treatment datasets Email and IST, with the results shown in Figure~\ref{fig:ablationstudy}.

From Figure~\ref{fig:ablationstudy}, we find that both w/o Adv and w/o Cov achieve lower $\Delta_{\mathrm{fair}}$ than Vanilla on Email and IST, indicating that the adversarial debiasing module and covariance-based fairness regularization term each contribute meaningfully to mitigating unfairness in treatment decisions. Across the four methods, the treatment values remain very similar, while FairITR attains the lowest $\Delta_{\mathrm{fair}}$ on both datasets. This shows that combining the two fairness components provides the most substantial fairness improvement with little loss in Value. In particular, the relative decrease in Value from Vanilla to FairITR is less than 1\% on both datasets, indicating a low treatment value cost for the fairness improvement. These results suggest that FairITR can greatly reduce fairness disparity at a negligible cost in treatment value, supporting the effectiveness of its design.

\subsection{Hyperparameter Sensitivity (RQ3)}
\begin{figure}[htbp]
  \centering
  \begin{subfigure}[t]{0.495\columnwidth}
    \centering
    \includegraphics[width=\linewidth]{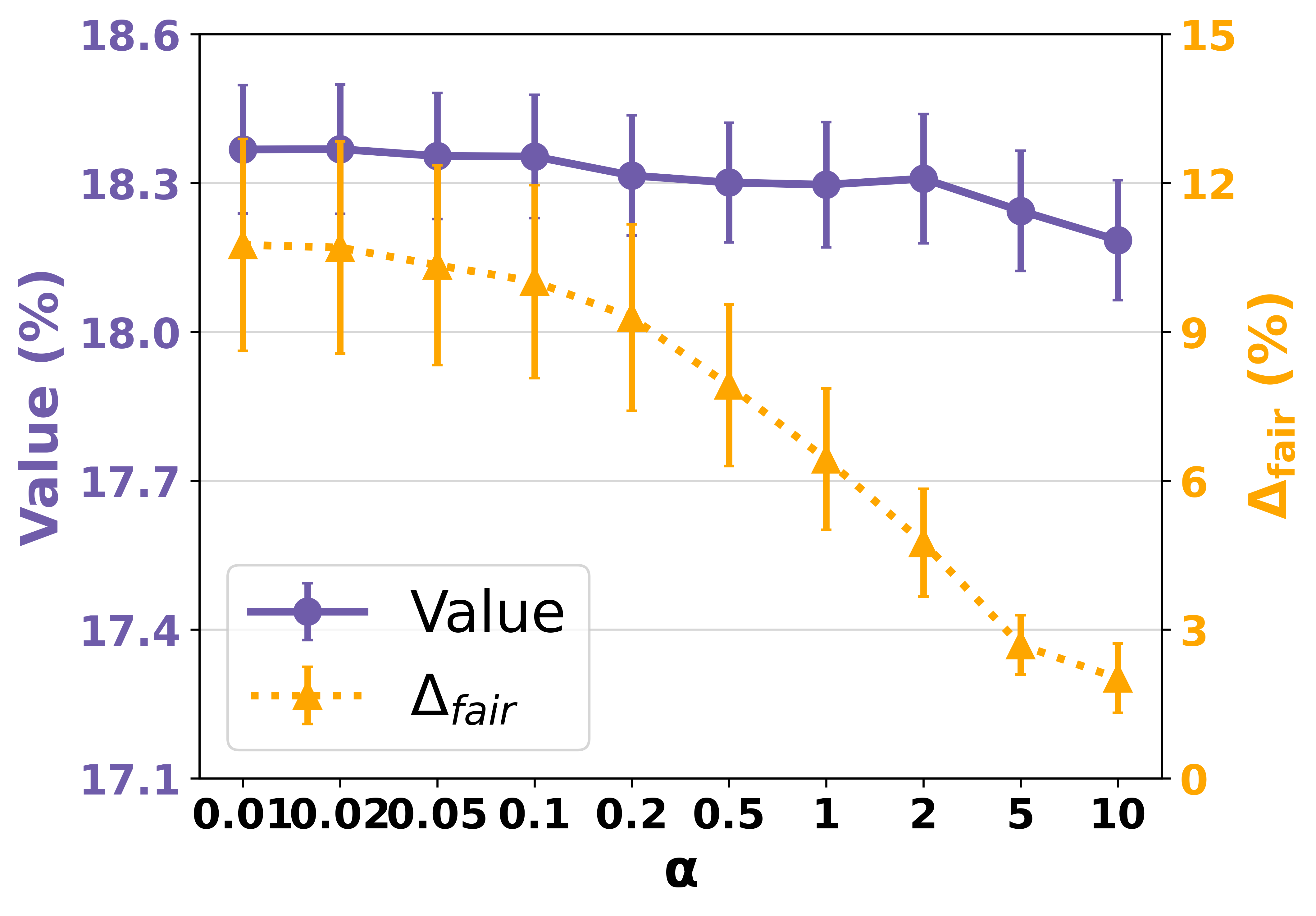}
    \caption{$\alpha$ for $\mathcal{L}_{\mathrm{cov}}$}
    \label{fig:sensitivity_alpha}
  \end{subfigure}
  \hfill 
  \begin{subfigure}[t]{0.495\columnwidth}
    \centering
    \includegraphics[width=\linewidth]{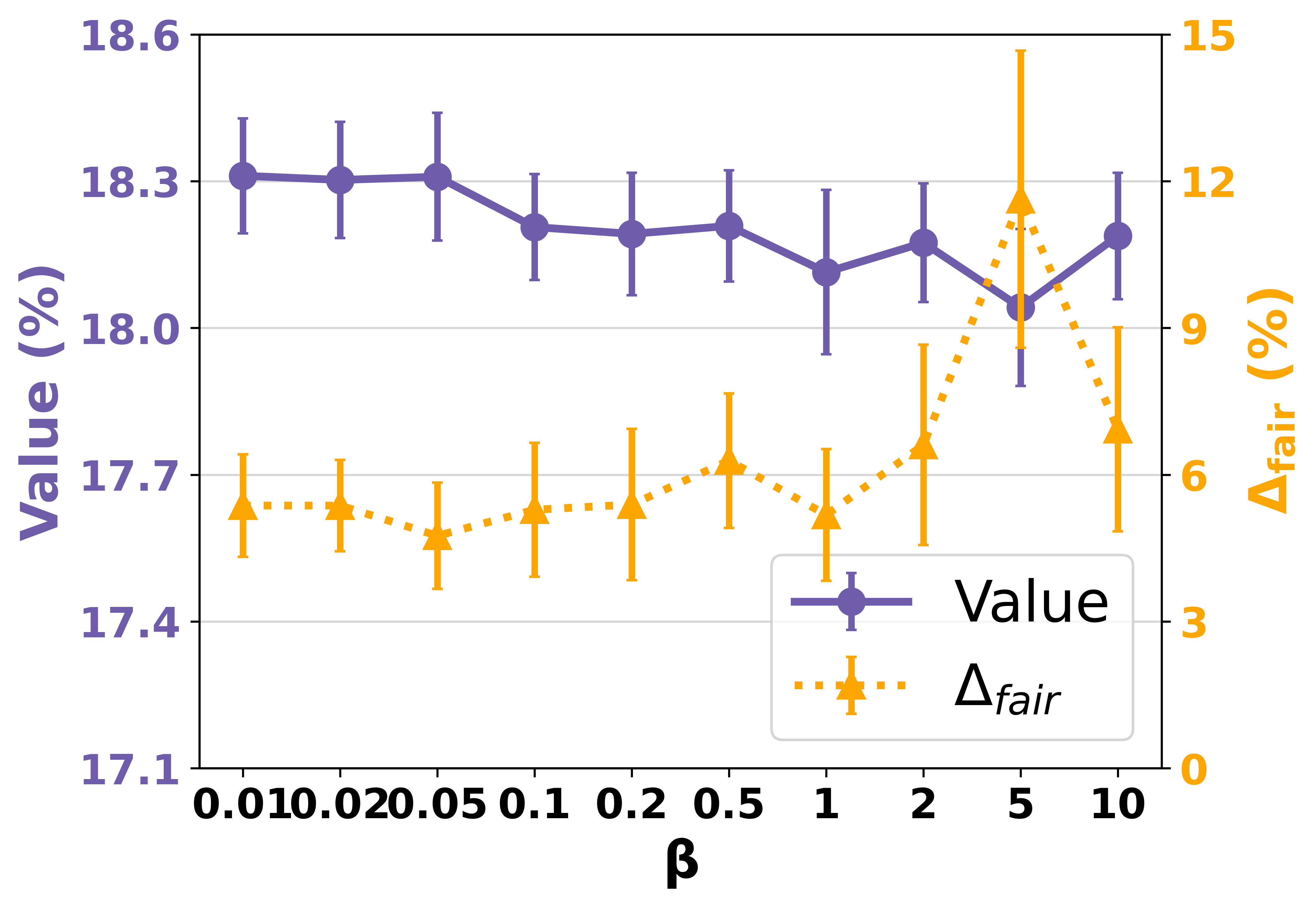}
    \caption{$\beta$ for $\mathcal{L}_{\mathrm{adv}}$}
    \label{fig:sensitivity_beta}
  \end{subfigure}
\caption{Hyperparameter sensitivity analysis of FairITR on the Email dataset. All values are shown on a percentage scale.}
\Description{Two line plots showing the sensitivity of FairITR to $\alpha$ and $\beta$ on the Email dataset. Each plot displays treatment value and fairness disparity across a range of hyperparameter values with standard error bars.}
\label{fig:sensitivityanalysis}
\end{figure}

FairITR involves two key hyperparameters, $\alpha$ and $\beta$, in its training objective \eqref{eqn:overallobjective}, which control the strengths of the covariance-based regularization and the adversarial debiasing, respectively. To answer RQ3, we analyze the sensitivity of FairITR to these hyperparameters. For demonstration, we perform the analysis using the Email dataset. The optimal values selected by grid search are $\alpha=2$ and $\beta=0.05$. We fix one hyperparameter at its selected value while varying the other over $\{0.01, 0.02, 0.05, 0.1, 0.2, 0.5, 1, 2, 5, 10\}$, and evaluate the resulting ITRs in terms of Value and $\Delta_{\mathrm{fair}}$. The results are shown in Figure~\ref{fig:sensitivityanalysis}. All other model specifications follow those in Section~\ref{subsubsec:implmentationdetails}.

Figure~\ref{fig:sensitivity_alpha} shows that $\Delta_{\mathrm{fair}}$ decreases steadily as $\alpha$ increases. It falls below $0.05$ at $\alpha=2$ and further decreases to approximately $0.02$ at $\alpha=10$. This improvement in fairness, however, comes with some loss in Value. The decline in Value is mild as $\alpha$ increases up to 2, but becomes more pronounced for larger values of $\alpha$.
Figure~\ref{fig:sensitivity_beta} reveals a less monotonic pattern with respect to $\beta$. Increasing $\beta$ initially reduces $\Delta_{\mathrm{fair}}$, which reaches its lowest level at $\beta=0.05$, but further increases in $\beta$ generally worsen fairness. One possible reason is that an overly large $\beta$ may amplify the instability of adversarial training and make the optimization of $g_{\mathrm{nn}}$ difficult. Therefore, the strength of adversarial debiasing should be carefully controlled. Meanwhile, Value remains relatively stable for small $\beta$ but tends to decline once $\beta$ exceeds $0.05$. Overall, the effect of $\alpha$ on FairITR performance is relatively smooth, whereas $\beta$ exhibits a more variable influence, reflecting the more complex dynamics of adversarial training. These results suggest that $\alpha=2$ and $\beta=0.05$ provide a favorable balance between treatment value and fairness for FairITR on Email.

\section{Conclusion}
Fairness concerns have become increasingly important when estimating ITRs. Meanwhile, practical decision-making problems often involve multiple treatment options, yet fairness-aware multicategory ITR methods remain underexplored. To bridge this gap, we introduce a fairness notion for multicategory ITRs and formulate the corresponding fair ITR learning problem. We further propose FairITR, a novel framework designed to mitigate unfairness in ITRs while maintaining high treatment value. Specifically, FairITR builds on angle-based classification for multicategory ITR learning and integrates adversarial debiasing with covariance-based fairness regularization to improve decision fairness. Extensive experiments on real-world datasets demonstrate that FairITR achieves favorable performance in both treatment value and fairness compared with existing baselines.
There are several interesting directions for future research. While the fairness definition considered in this work focuses on disparities in treatment decisions, future studies could extend fairness considerations to broader social welfare outcomes induced by these decisions. Other bias mitigation techniques could also be explored and incorporated into fair ITR learning to further enhance the flexibility and effectiveness of the framework.

\appendix

\section{Proof of Proposition~\ref{thm:advlearning}} \label{appendixsec:proof}

\begin{proof}
    Following Proposition 1 of \citet{goodfellow2014generative}, for a fixed $g_{\mathrm{hid}}$, the optimal adversary that maximizes $\Psi(g_{\mathrm{hid}}, g_{\mathrm{adv}})$ is
    \begin{equation*}
        g_{\mathrm{adv}}^*(\mathbf{h}) = \frac{f(\mathbf{h} \mid S=1)}{f(\mathbf{h} \mid S=1) + f(\mathbf{h} \mid S=0)}.
    \end{equation*}
    Substituting $g_{\mathrm{adv}}^*$ into $\Psi(g_{\mathrm{hid}}, g_{\mathrm{adv}})$ gives
    \begin{align*}
        \Psi(g_{\mathrm{hid}}, g_{\mathrm{adv}}^*) =&\; \mathrm{E}_{\mathbf{h} \sim f(\mathbf{h} \mid S=1)} \left[ \log \frac{f(\mathbf{h} \mid S=1)}{f(\mathbf{h} \mid S=1) + f(\mathbf{h} \mid S=0)} \right] \\
        &+  \mathrm{E}_{\mathbf{h} \sim f(\mathbf{h} \mid S=0)} \left[ \log \frac{f(\mathbf{h} \mid S=0)}{f(\mathbf{h} \mid S=1) + f(\mathbf{h} \mid S=0)} \right] \\
        =&\; -\log(4) + 2 \mathrm{JSD}(f(\mathbf{h} \mid S=1) \| f(\mathbf{h} \mid S=0)),
    \end{align*}
    where $\mathrm{JSD}(\cdot \| \cdot)$ denotes the Jensen–Shannon divergence. Since this divergence is nonnegative and equals zero if and only if the two distributions are identical, $\Psi(g_{\mathrm{hid}}, g_{\mathrm{adv}}^*)$ attains its global minimum if and only if $f(\mathbf{h} \mid S=1) = f(\mathbf{h} \mid S=0)$ almost everywhere. That is, $\mathbf{h} \independent S$. Now for any output layer $g_{\mathrm{out}}: \mathbb{R}^q \rightarrow \mathbb{R}^{K-1}$, since $g_{\mathrm{out}}(\mathbf{h})$ is a deterministic function of $\mathbf{h}$, the independence between $\mathbf{h}$ and $S$ implies $g_{\mathrm{out}}(\mathbf{h}) \independent S$. The resulting treatment decision is
    \begin{equation*}
        d(\mathbf{X}, S) = \argmax_{k \in \{1,\dots,K\}} \mathbf{w}_k^T g_{\mathrm{out}}(\mathbf{h}).
    \end{equation*}
    Assuming ties are resolved by a fixed deterministic rule, $d(\mathbf{X}, S)$ is a deterministic function of $g_{\mathrm{out}}(\mathbf{h})$. Hence, $d(\mathbf{X}, S) \independent S$, and this completes the proof.
\end{proof}

\section{Additional Implementation Details} \label{appendixsec:experimentdetails}
This section presents supplementary information on the numerical experiments. For the propensity scores $\pi_A(\mathbf{X}, S)$, we estimate them in the RHC dataset (an observational study) by fitting a logistic regression model of $A$ on $(\mathbf{X}, S)$ using the training set. For the randomized Email and IST datasets, the treatment assignment probabilities are known by design and equal to $1/3$ and $1/4$ for each treatment option, respectively. We next provide further details on the implementation of each baseline introduced in Section~\ref{subsubsec:baselines}.
\begin{itemize}[leftmargin=*]
    \item $l_1$-PLS: Following \citet{zhao2012estimating}, we use $\{1, (\mathbf{X}, S), A, (\mathbf{X}, S)A\}$ as the basis function set in the outcome regression. The LASSO regularization parameter is selected using 10-fold cross-validation.
    \item OWL: We implement this method using the \texttt{owl} function in the R package \texttt{DynTxRegime} \cite{holloway2025dyntxregime}, with all function arguments set to their default values.
    \item OWL-MD: For this method, we use the open-source implementation code released by \citet{huang2019multicategory}.
    \item OML and MOML: Both methods are implemented using the \texttt{moml} function in the R package \texttt{abclass} \cite{wang2022abclass}. We adopt the LUM loss defined in \eqref{eqn:lumloss}, with $\zeta=1$ and $\eta=0$, which are the default values in the \texttt{moml} function. This also matches the specification used in FairITR. We use LASSO regularization to mitigate overfitting, and the penalty strength is determined using 10-fold cross-validation.
    \item UF-ITR: We use the implementation code from the GitHub repository provided in \citet{lee2024effective}. Since UF-ITR includes a variable screening step that selects important covariates for the outcome before ITR estimation, we use Ridge regularization for the subsequent ITR fitting instead of LASSO regularization, given the small to moderate number of covariates in our datasets.
    \item FairPol: We implement it using the open-source code provided by \citet{frauen2024fair}. FairPol has a two-stage design, with a representation network that maps $(\mathbf{X}, S)$ to fair representations and a policy network that learns the ITR from these representations. The representation network is trained adversarially, with both the outcome prediction and sensitive attribute networks implemented as linear models. To adapt FairPol to our deterministic ITR setting, we build its policy network using the ITR module in Section~\ref{sec:ITRModule}. For a fair comparison with FairITR, we match the architectures of the representation and policy networks to those of $g_{\mathrm{hid}}$ and $g_{\mathrm{out}}$, respectively. The representation loss includes a hyperparameter that controls the trade-off between preserving outcome information and removing sensitive information. We select its value by cross-validation with a grid search, resulting in values of $0.2$, $0.001$, and $5$ for RHC, Email, and IST, respectively.
\end{itemize}
For all baselines except FairPol, their FTU variants follow the same implementation, with the model input changed from $(\mathbf{X}, S)$ to $\mathbf{X}$.

As for FairITR, we provide additional details on its optimization setup and hyperparameter selection. For the Adam optimizers used during training, the learning rate and weight decay are set to $10^{-4}$ and $5 \times 10^{-3}$ for RHC, $5 \times 10^{-4}$ and $10^{-5}$ for Email, and $10^{-4}$ and $10^{-5}$ for IST, respectively. For the grid search of $\alpha$ and $\beta$ based on cross-validation, we vary each over 
\begin{equation*}
    \{0.001, 0.005, 0.01, 0.02, 0.05, 0.1, 0.2, 0.5, 1, 2, 5, 10, 20, 50\},
\end{equation*}
and the selected values are $(\alpha, \beta) = (10, 0.5)$ for RHC, $(2, 0.05)$ for Email, and $(20, 0.5)$ for IST.

\section*{GenAI Usage Disclosure}
The authors acknowledge the use of ChatGPT-5.6 solely for grammar checking, language polishing to improve conciseness, and table formatting during manuscript preparation. No generative AI tools were used to generate scientific content.

\bibliographystyle{ACM-Reference-Format}
\bibliography{reference}


\end{document}